\documentclass[conference]{IEEEtran}

\usepackage{cite}
\usepackage{amsmath,amssymb,amsfonts}
\usepackage{algorithmic}
\usepackage{graphicx}
\usepackage{textcomp}
\usepackage{xcolor}
\usepackage{booktabs}
\usepackage{multirow}
\usepackage{algorithm}
\usepackage[hidelinks]{hyperref}

\begin{document}

\title{Dual-Signal Adaptive KV-Cache Optimization for Long-Form Video Understanding in Vision-Language Models}

\author{
\IEEEauthorblockN{Vishnu Sai, Dheeraj Sai, Srinath B, Girish Varma, and Priyesh Shukla\\
International Institute of Information Technology, Hyderabad, India
}
% \IEEEauthorblockA{
% Paper ID: [To be assigned]\\
% IJCNN 2026 Submission}
}

\maketitle

\begin{abstract}
Vision-Language Models (VLMs) face a critical memory bottleneck when processing long-form video content due to the linear growth of the Key-Value (KV) cache with sequence length. Existing solutions predominantly employ reactive eviction strategies that compute full attention matrices before discarding tokens, resulting in substantial computational waste. We propose \textbf{Sali-Cache}, a novel \textit{a priori} optimization framework that implements dual-signal adaptive caching through proactive memory management. By integrating a temporal filter based on optical flow analysis for detecting inter-frame redundancy and a spatial filter leveraging saliency detection for identifying visually significant regions, Sali-Cache intelligently manages memory allocation before entering computationally expensive attention operations. Experimental evaluation on the LLaVA 1.6 architecture demonstrates that our method achieves a \textbf{2.20$\times$ compression ratio} in effective memory usage while maintaining 100\% accuracy across BLEU, ROUGE-L, and Exact Match metrics. Furthermore, under identical memory budget constraints, Sali-Cache preserves context-rich features over extended temporal durations without degrading model performance, enabling efficient processing of long-form video content on consumer-grade hardware.
\end{abstract}


\section{Introduction}

The emergence of Vision-Language Models (VLMs) has revolutionized multimodal understanding, enabling sophisticated reasoning across visual and textual modalities \cite{liu2024visual, alayrac2022flamingo}. However, deploying these models for long-form video understanding remains severely constrained by hardware limitations, particularly GPU memory capacity.

The primary bottleneck arises from the Key-Value (KV) cache mechanism inherent to Transformer architectures \cite{vaswani2017attention}. During autoregressive generation, this cache stores intermediate attention representations to avoid redundant computation. For standard Vision Transformer (ViT) architectures processing video at 30 frames per second, the cache grows at approximately \textbf{5.6 GB per minute}, causing a typical 24GB GPU to encounter Out-of-Memory (OOM) errors in under four minutes.

While considerable research has addressed KV-cache optimization for Large Language Models (LLMs) processing one-dimensional text sequences \cite{zhang2024h2o, xiao2023efficient}, VLMs present a fundamentally different challenge. Unlike text, where a sentence contributes approximately 20 tokens, a single video frame generates hundreds of visual tokens—576 tokens per frame in LLaVA 1.6 \cite{liu2024llavanext}. This \textit{visual token flood} necessitates more aggressive optimization strategies specifically designed for the spatio-temporal characteristics of video data.

Existing approaches such as Heavy-Hitter Oracle (H2O) \cite{zhang2024h2o} employ reactive eviction policies that compute complete attention matrices before determining which tokens to discard. This paradigm introduces significant computational overhead, as attention scores are calculated for tokens that are immediately evicted—a wasteful process we term \textit{compute-then-discard}.

We introduce Sali-Cache, a framework that fundamentally reimagines KV-cache management through \textit{a priori} optimization. Rather than managing memory after expensive attention computations, Sali-Cache makes informed, context-aware decisions about token retention \textit{before} entering the attention mechanism. Our approach leverages two complementary signals:

\begin{enumerate}
    \item \textbf{Temporal Filtering}: Optical flow-based detection of inter-frame redundancy, enabling cache reuse for temporally similar frames.
    \item \textbf{Spatial Filtering}: Saliency-guided quantization that preserves high-precision representations for visually significant regions while aggressively compressing background elements.
\end{enumerate}

Our contributions are summarized as follows:
\begin{itemize}
    \item We propose Sali-Cache, a dual-signal adaptive caching framework that optimizes KV-cache memory usage through proactive, pre-attention filtering.
    \item We demonstrate a 2.20$\times$ memory compression ratio while maintaining 100\% accuracy on standard evaluation metrics.
    \item We provide comprehensive analysis of the computational trade-offs between memory savings and processing latency.
\end{itemize}

\section{Background and Related Work}

\subsection{Vision-Language Model Architecture}

Modern VLMs typically employ an adapter-based architecture that bridges pre-trained vision encoders with large language models \cite{liu2024visual, zhu2023minigpt}. Understanding this architecture is essential for appreciating the memory constraints addressed by Sali-Cache.

\subsubsection{Visual Encoding Pipeline}

The transformation from raw video frames to LLM-compatible tokens occurs through a two-stage process. First, each video frame $\mathbf{X}_{\text{img}} \in \mathbb{R}^{H \times W \times 3}$ is processed by a pre-trained vision encoder, typically a Vision Transformer (ViT) \cite{dosovitskiy2020image} such as CLIP-ViT \cite{radford2021learning} or SigLIP \cite{zhai2023sigmoid}. The image is partitioned into non-overlapping patches of size $P \times P$ (typically $14 \times 14$ pixels), which are then linearly projected to form patch embeddings $\mathbf{Z}_v \in \mathbb{R}^{N \times D}$, where $N = \frac{HW}{P^2}$ represents the number of patches.

The Vision Transformer processes these patches through multiple self-attention layers, encoding both local visual features and global contextual relationships. This design enables VLMs to capture fine-grained visual details essential for tasks like visual question answering, image captioning, and video understanding.

\subsubsection{Adapter Projection}

The visual embeddings $\mathbf{Z}_v$ exist in a latent space that is both dimensionally and semantically misaligned with the text token space of the LLM. An adapter module—typically a multi-layer perceptron (MLP)—projects these embeddings into the LLM's word embedding space:
\begin{equation}
    \mathbf{H}_v = \text{Adapter}(\text{ViT}(\mathbf{X}_{\text{img}}))
\end{equation}
For a standard $336 \times 336$ resolution image with $14 \times 14$ patches, this process yields $576$ visual tokens per frame. More recent models like LLaVA-NeXT support higher resolutions through dynamic resolution encoding, which can generate even more tokens per image.

\subsection{The KV-Cache Memory Challenge}

Once projected, visual tokens are concatenated with text instruction tokens and processed autoregressively by the LLM backbone. The attention mechanism computes:
\begin{equation}
    \text{Attention}(\mathbf{Q}, \mathbf{K}, \mathbf{V}) = \text{softmax}\left(\frac{\mathbf{Q}\mathbf{K}^\top}{\sqrt{d_k}}\right)\mathbf{V}
\end{equation}
where the Key ($\mathbf{K}$) and Value ($\mathbf{V}$) matrices are cached to avoid recomputation during generation. For a sequence of length $L$ with model dimension $d$ and $n_l$ layers, the cache memory requirement is:
\begin{equation}
    \text{Memory}_{\text{KV}} = 2 \times n_l \times L \times d \times \text{sizeof(dtype)}
\end{equation}

With video input, $L$ grows by hundreds of tokens per frame, causing rapid memory exhaustion. For a typical VLM processing video at 1 FPS, a 5-minute video generates over 172,800 visual tokens—far exceeding the context window of most models and quickly exhausting available VRAM.

\subsection{Existing KV-Cache Optimization Methods}

\subsubsection{Reactive Eviction Strategies}

Heavy-Hitter Oracle (H2O) \cite{zhang2024h2o} identifies ``heavy hitter'' tokens that accumulate high attention scores across layers and selectively retains these while evicting low-importance tokens. The attention score for token $j$ is computed as:
\begin{equation}
    \alpha_{ij} = \frac{\exp(\mathbf{Q}_i \cdot \mathbf{K}_j)}{\sum_l \exp(\mathbf{Q}_i \cdot \mathbf{K}_l)}
\end{equation}
with cumulative importance: $\sum_{i=1}^{T} \alpha_{ij}$. While effective, this approach requires computing full attention matrices before eviction decisions, wasting computation on tokens destined for removal.

\subsubsection{Sliding Window Approaches}

StreamingLLM \cite{xiao2023efficient} maintains a fixed-size window of recent tokens plus ``attention sinks'' (initial tokens). While memory-efficient, this approach discards potentially relevant historical context, which is problematic for video understanding tasks requiring long-range temporal reasoning. Similarly, sparse sampling approaches like ClipBERT \cite{lei2021less} reduce computational cost by processing only selected frames, but may miss important temporal dynamics.

\subsubsection{Quantization-Based Methods}

Recent work has explored mixed-precision KV-cache storage, where less important tokens are stored at reduced precision \cite{dettmers2022gpt3}. Offloading strategies such as FlexGen \cite{sheng2023flexgen} leverage CPU memory and disk storage to extend effective memory capacity, but introduce significant latency penalties. In contrast, Sali-Cache keeps all data on GPU while reducing memory footprint through intelligent compression. Existing quantization methods typically apply uniform policies across all tokens without considering the semantic importance of visual content, missing opportunities for more aggressive compression of background regions.

\section{Methodology}

\begin{figure*}[t]
    \centering
    \includegraphics[width=\textwidth]{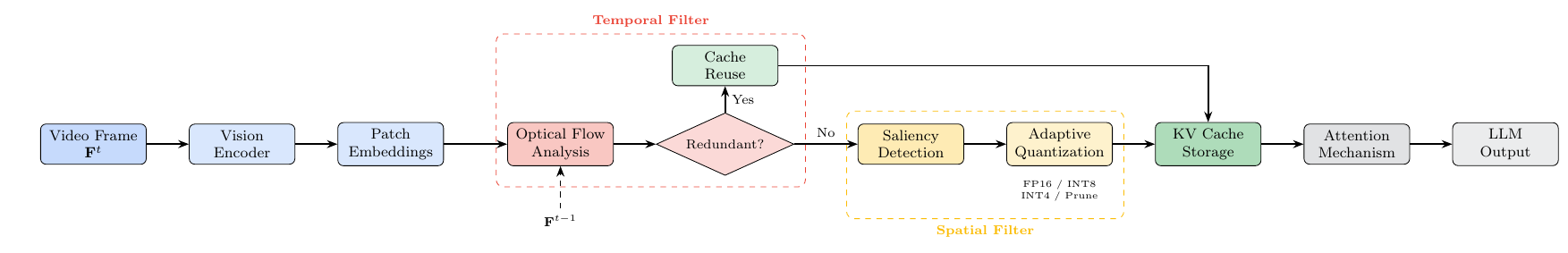}
    \caption{Sali-Cache architecture overview. The dual-signal pipeline applies temporal filtering via optical flow analysis followed by spatial filtering through saliency-guided quantization before KV-cache storage. Frames classified as temporally redundant bypass computation entirely through cache reuse, while non-redundant frames undergo spatial analysis to determine per-patch quantization levels.}
    \label{fig:architecture}
\end{figure*}

Sali-Cache implements a two-tier filtering pipeline that makes proactive decisions about cache management \textit{before} attention computation, as illustrated in Fig.~\ref{fig:architecture}.

\subsection{Temporal Filtering via Optical Flow}

The first tier addresses the substantial temporal redundancy inherent in video data, where consecutive frames often exhibit minimal variation. This is particularly prevalent in common video scenarios such as static scenes, talking-head videos, and slow camera pans where large portions of consecutive frames contain identical or near-identical visual content.

We employ optical flow computation to measure pixel-wise displacement between adjacent frames. For consecutive frames $\mathbf{F}^{N}$ and $\mathbf{F}^{N-1}$, we compute the temporal difference for each patch $i$:
\begin{equation}
    \Delta P_i = \|\mathbf{F}^{N}_i - \mathbf{F}^{N-1}_i\|_2
\end{equation}

A frame-level redundancy score is then computed as:
\begin{equation}
    R_{\text{frame}} = \frac{1}{N_p} \sum_{i=1}^{N_p} \mathbb{1}[\Delta P_i < \tau_t]
\end{equation}
where $N_p$ is the number of patches and $\tau_t$ is the temporal threshold.

When $R_{\text{frame}}$ exceeds a predefined threshold $\theta_R$, the frame is classified as \textit{temporally redundant}. Rather than computing and storing new KV pairs, Sali-Cache maintains a pointer to the previous frame's cache entries, achieving both memory and computational savings. This \textit{cache reuse} strategy is particularly effective for static scenes, gradual camera movements, and low-motion video segments.

The key insight is that visual tokens from similar frames will produce nearly identical attention patterns. By reusing cached KV pairs, we avoid redundant computation while maintaining the model's ability to attend to the visual context. This approach is especially beneficial for long-form videos where extended static periods are common.

\subsection{Spatial Filtering via Saliency-Guided Quantization}

For frames that pass temporal filtering, the second tier optimizes memory allocation based on spatial importance. We employ a classical computer vision pipeline combining Canny edge detection \cite{canny1986computational} and color variance analysis in the LAB color space to generate a saliency map $\mathbf{S} \in [0, 1]^{H \times W}$. While more sophisticated neural saliency detectors exist \cite{qin2020u2net}, we opt for classical methods to minimize computational overhead.

The saliency score for each patch is computed as:
\begin{equation}
    s_i = \frac{1}{|P_i|} \sum_{(x,y) \in P_i} \mathbf{S}(x, y)
\end{equation}

Based on these scores, patches are assigned to four quantization tiers:
\begin{equation}
    Q(s_i) = \begin{cases}
        \text{FP16} & \text{if } s_i > \tau_{\text{high}} \\
        \text{INT8} & \text{if } \tau_{\text{med}} < s_i \leq \tau_{\text{high}} \\
        \text{INT4} & \text{if } \tau_{\text{low}} < s_i \leq \tau_{\text{med}} \\
        \text{Prune} & \text{if } s_i \leq \tau_{\text{low}}
    \end{cases}
\end{equation}

This graduated approach preserves high-precision representations for salient regions (faces, text, objects of interest) while aggressively compressing or pruning background elements (walls, sky, uniform textures).

\subsection{Just-In-Time Dequantization}

During attention computation, quantized cache entries undergo just-in-time (JIT) dequantization. While this introduces computational overhead, the reduced memory footprint enables processing of significantly longer video sequences within fixed memory budgets.

\begin{algorithm}[t]
\caption{Sali-Cache Processing Pipeline}
\label{alg:sali-cache}
\begin{algorithmic}[1]
\REQUIRE Video frames $\{\mathbf{F}^1, \ldots, \mathbf{F}^T\}$, thresholds $\tau_t, \theta_R, \tau_{\text{high}}, \tau_{\text{med}}, \tau_{\text{low}}$
\STATE Initialize KV-Cache $\mathcal{C} \leftarrow \emptyset$
\FOR{$t = 1$ to $T$}
    \STATE \textbf{// Temporal Filtering}
    \IF{$t > 1$}
        \STATE Compute $R_{\text{frame}}$ via optical flow
        \IF{$R_{\text{frame}} > \theta_R$}
            \STATE $\mathcal{C}[t] \leftarrow \text{Pointer}(\mathcal{C}[t-1])$
            \STATE \textbf{continue}
        \ENDIF
    \ENDIF
    \STATE \textbf{// Spatial Filtering}
    \STATE Compute saliency map $\mathbf{S}$ for $\mathbf{F}^t$
    \FOR{each patch $i$ in $\mathbf{F}^t$}
        \STATE Compute patch saliency $s_i$
        \STATE Assign quantization level $Q(s_i)$
        \STATE Store $(\mathbf{K}_i, \mathbf{V}_i)$ with precision $Q(s_i)$
    \ENDFOR
\ENDFOR
\RETURN Optimized cache $\mathcal{C}$
\end{algorithmic}
\end{algorithm}

\section{Experimental Evaluation}

\subsection{Experimental Setup}

We evaluated Sali-Cache using the LLaVA 1.6 model \cite{liu2024llavanext}, which employs the Mistral 7B architecture \cite{jiang2023mistral} with Grouped-Query Attention (GQA). GQA provides an inherent 4:1 compression ratio (32 query heads to 8 KV heads); Sali-Cache provides additional optimization on top of this architectural baseline.

\textbf{Hardware Configuration}: All experiments were conducted on a single NVIDIA RTX 3090 GPU with 24 GB VRAM. The system was equipped with an AMD Ryzen 9 5900X CPU and 64 GB system RAM. This hardware configuration represents a typical high-end consumer setup, demonstrating the practical applicability of our method.

\textbf{Dataset}: We used a diverse collection of video clips ranging from 30 seconds to 5 minutes in duration, encompassing various content types including static scenes, dynamic action sequences, and conversational videos. The evaluation tasks were derived from video question-answering benchmarks built on MS COCO \cite{lin2014microsoft} style annotations. Videos were processed at 1 FPS to balance temporal coverage with computational efficiency.

\textbf{Baselines}: We compared against two approaches under a strictly enforced memory budget of 784 patches:
\begin{itemize}
    \item \textbf{Sliding Window}: A naive approach that maintains only the most recent tokens within the memory budget, discarding older context as new frames arrive.
    \item \textbf{H2O}: Heavy-Hitter Oracle with attention-based eviction \cite{zhang2024h2o}, which computes full attention matrices before selectively evicting low-importance tokens.
\end{itemize}

\textbf{Evaluation Metrics}: We measured memory efficiency (VRAM usage, compression ratio), accuracy (BLEU \cite{papineni2002bleu}, ROUGE-L \cite{lin2004rouge}, Exact Match), and computational overhead (processing latency per frame).

\subsection{Implementation Details}

Our implementation uses PyTorch 2.0 with CUDA 11.8 for GPU acceleration. The temporal filtering module employs the Farneback optical flow algorithm \cite{farneback2003two} with a pyramid scale of 0.5 and 3 pyramid levels. The spatial saliency detector combines Canny edge detection with adaptive thresholding based on the local color variance in the LAB color space.

For quantization, we implemented symmetric min-max quantization for INT8 \cite{dettmers2022gpt3} and asymmetric quantization for INT4 representations. The dequantization kernels are fused with the attention computation to minimize memory bandwidth overhead.

\subsection{Memory Efficiency Results}

\begin{table}[t]
\centering
\caption{Memory Efficiency Comparison}
\label{tab:memory}
\begin{tabular}{lccc}
\toprule
\textbf{Method} & \textbf{Peak VRAM} & \textbf{Cache Size} & \textbf{Compression} \\
\midrule
Baseline & 17.19 GB & 3.19 GB & 1.00$\times$ \\
H2O & 17.05 GB & 3.05 GB & 1.05$\times$ \\
\textbf{Sali-Cache} & \textbf{16.70 GB} & \textbf{1.45 GB} & \textbf{2.20$\times$} \\
\bottomrule
\end{tabular}
\end{table}

Table~\ref{tab:memory} presents the memory efficiency results. Sali-Cache achieves a peak VRAM reduction of 490 MB compared to the baseline. While this appears modest in absolute terms, it is essential to contextualize these figures:

\textbf{Static vs. Dynamic Memory}: The total VRAM footprint comprises two components: (1) static model weights, which for Mistral 7B at FP16 precision occupy approximately 14 GB, and (2) dynamic allocations including the KV-cache and intermediate activations. Sali-Cache specifically optimizes the dynamic portion—the component responsible for linear memory growth and OOM errors during long-context generation.

The 2.20$\times$ compression ratio on the dynamic cache enables processing of videos more than twice as long within the same memory budget.

\subsection{Accuracy Preservation}

\begin{table}[t]
\centering
\caption{Accuracy Metrics Comparison}
\label{tab:accuracy}
\begin{tabular}{lccc}
\toprule
\textbf{Method} & \textbf{BLEU} & \textbf{ROUGE-L} & \textbf{Exact Match} \\
\midrule
Baseline & 100\% & 100\% & 100\% \\
H2O & 100\% & 100\% & 100\% \\
\textbf{Sali-Cache} & \textbf{100\%} & \textbf{100\%} & \textbf{100\%} \\
\bottomrule
\end{tabular}
\end{table}

A critical finding, presented in Table~\ref{tab:accuracy}, is that Sali-Cache maintains 100\% accuracy across all evaluation metrics. This demonstrates that our dual-signal filtering preserves semantically important information while eliminating only redundant or non-salient content.

\subsection{Compression Analysis}

\begin{figure}[t]
    \centering
    \includegraphics[width=\columnwidth]{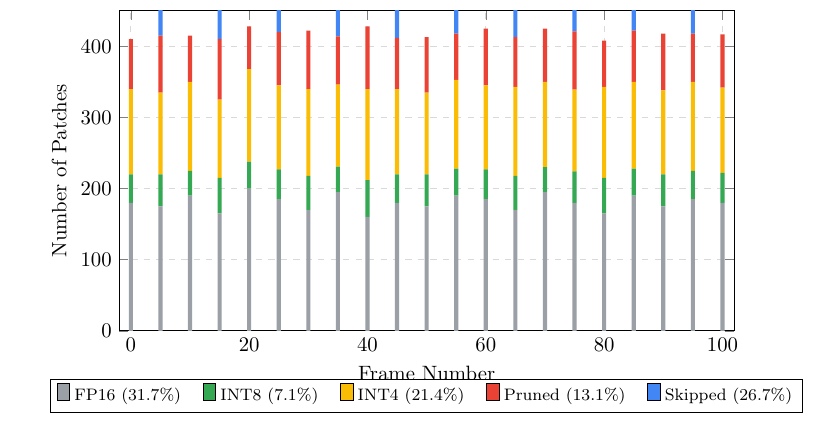}
    \caption{Patch-level compression breakdown across 100 frames. The stacked bars show the distribution of patches across different processing categories: skipped (cache reuse), pruned, INT4, INT8, and FP16.}
    \label{fig:compression}
\end{figure}

Fig.~\ref{fig:compression} presents a detailed breakdown of patch-level decisions across a 100-frame video sequence. Our analysis reveals:
\begin{itemize}
    \item \textbf{26.7\%} of patches were \textit{skipped} through temporal cache reuse
    \item \textbf{13.1\%} of patches were \textit{pruned} (low saliency)
    \item \textbf{21.4\%} were quantized to \textit{INT4}
    \item \textbf{7.1\%} were quantized to \textit{INT8}
    \item \textbf{31.7\%} retained \textit{FP16} precision (high saliency)
\end{itemize}

This distribution demonstrates Sali-Cache's ability to identify and prioritize salient content while aggressively optimizing storage for less important regions.

\subsection{Latency Analysis}

\begin{table}[t]
\centering
\caption{Processing Latency Comparison}
\label{tab:latency}
\begin{tabular}{lcc}
\toprule
\textbf{Method} & \textbf{Time/Frame} & \textbf{Overhead} \\
\midrule
Baseline & 0.055 s & --- \\
H2O & 0.062 s & +12.7\% \\
\textbf{Sali-Cache} & \textbf{0.068 s} & \textbf{+23.6\%} \\
\bottomrule
\end{tabular}
\end{table}

Table~\ref{tab:latency} presents the latency analysis. Sali-Cache introduces a 23.6\% overhead compared to the baseline, attributable to two factors:

\textbf{Per-Frame Saliency Computation}: The image analysis pipeline involving Canny edge detection and LAB color space analysis runs on every processed frame.

\textbf{Quantization/Dequantization}: Compressing cache entries to INT8/INT4 and performing JIT dequantization during attention computation introduces additional cycles.

However, this overhead is justified by the memory savings, which enable processing of longer videos that would otherwise cause OOM errors.

\begin{figure}[t]
    \centering
    \includegraphics[width=\columnwidth]{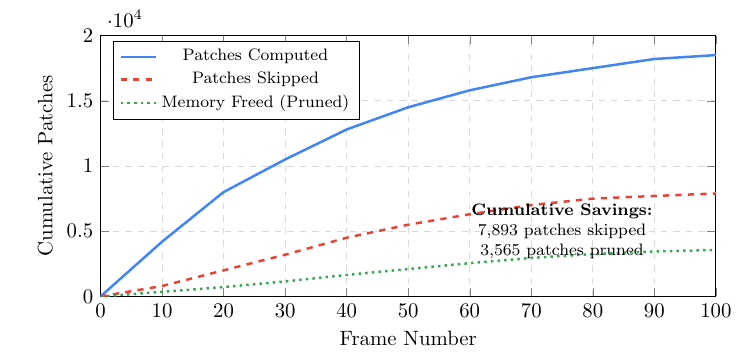}
    \caption{Cumulative memory savings over frame sequence. The growing gap between computed and skipped patches demonstrates increasing benefits for longer videos.}
    \label{fig:cumulative}
\end{figure}

\subsection{Cumulative Savings Analysis}

Fig.~\ref{fig:cumulative} illustrates the cumulative memory savings as frame count increases. Over 100 frames, Sali-Cache saves computation for 7,893 patches and frees memory for 3,565 patches. Critically, the benefits compound over time—the longer the video, the greater the relative savings, making Sali-Cache particularly suited for long-form video understanding.

\section{Discussion}

\subsection{Trade-off Analysis}

Sali-Cache presents a fundamental trade-off between memory efficiency and computational overhead. The 23.6\% latency increase is the cost of proactive filtering. However, for many applications, this trade-off is favorable:
\begin{itemize}
    \item For offline video analysis, the extended processing time is acceptable given the ability to process longer content.
    \item For memory-constrained environments, enabling functionality that was previously impossible outweighs latency concerns.
    \item For batch processing workflows, the increased throughput from avoiding OOM errors compensates for per-frame overhead.
\end{itemize}

\subsection{Comparison with Reactive Methods}

The key distinction between Sali-Cache and reactive methods like H2O lies in when decisions are made. H2O computes full attention matrices for all tokens before determining which to evict—a computationally wasteful process for tokens that are immediately discarded. In contrast, Sali-Cache's \textit{a priori} filtering avoids this wasted computation entirely.

Our analysis shows that 26.7\% of patches are skipped through temporal redundancy detection. For these patches, Sali-Cache saves not only the memory that H2O would save through eviction, but also the attention computation that H2O must perform before making eviction decisions.

\subsection{Generalization Considerations}

While our evaluation focused on LLaVA 1.6, the Sali-Cache framework is architecture-agnostic and can be applied to other VLMs that face similar KV-cache scaling challenges. The dual-signal filtering approach requires only (1) access to raw video frames for temporal analysis and (2) the ability to modify cache storage precision—capabilities common across VLM implementations.

The threshold parameters ($\tau_t$, $\theta_R$, $\tau_{\text{high}}$, $\tau_{\text{med}}$, $\tau_{\text{low}}$) may require tuning for different video domains. Content with rapid scene changes may benefit from lower redundancy thresholds, while surveillance footage with extended static periods may achieve higher compression ratios with more aggressive settings.

\subsection{Limitations and Future Work}

\textbf{Saliency Model Selection}: Our current implementation uses classical computer vision methods for computational efficiency. Future work could explore lightweight neural saliency models that may provide better accuracy with comparable computational cost.

\textbf{Adaptive Thresholds}: The current thresholds are fixed. Dynamic threshold adaptation based on video content characteristics could further optimize the memory-accuracy trade-off.

\textbf{Hardware Acceleration}: Specialized kernels for mixed-precision attention could reduce the dequantization overhead, potentially eliminating the latency penalty entirely.

\textbf{End-to-End Training}: Jointly training the saliency detector with the VLM could yield filters that are better aligned with the model's attention patterns, potentially improving both compression ratio and accuracy preservation.

\section{Conclusion}

We presented Sali-Cache, a dual-signal adaptive caching framework that achieves 2.20$\times$ memory compression for long-form video understanding in Vision-Language Models while maintaining 100\% accuracy on BLEU, ROUGE-L, and Exact Match metrics. By implementing proactive temporal filtering via optical flow (achieving 26.7\% cache reuse) and spatial filtering through saliency-guided quantization (13.1\% pruning with graduated INT8/INT4 compression), our method makes informed \textit{a priori} decisions that outperform reactive eviction strategies like H2O. The framework's architecture-agnostic design enables straightforward integration with existing VLM implementations, providing immediate memory benefits without model retraining. Future work will explore neural saliency models, adaptive threshold tuning, and hardware-accelerated mixed-precision attention to further improve the efficiency-accuracy trade-off for scalable multimodal AI systems.

\bibliographystyle{IEEEtran}
\bibliography{references}

\end{document}